# Design and implementation of a wireless instrument adapter

Kaori V. Laino[1], Thore Saathoff[1], Thiusius R. Savarimuthu[2],
Kim Lindberg Schwaner[2], Nils Gessert[1], Alexander Schlaefer[1]

[1] Hamburg University of Technology, Hamburg, Germany
[2] University of Southern Denmark, Odense, Denmark

Kontakt: kaori.laino@hotmail.de

## Abstract

The evaluation of new methods for control and manipulation in minimally invasive robotic surgery requires a handy and realistic setup. To decouple the investigation of new methods in active research fields from overall clinical systems, we propose an instrument adapter for the S line EndoWrist© instruments of the da Vinci© surgical system. The adapter is small and lightweight and can be mounted to any robot to mimic motion. We describe its design and implementation, as well as a setup to calibrate instruments to study precise motion control with regard to research fields like autonomy of surgical sub-tasks or soft tissue interaction. Our results indicate that each instrument requires individual calibration. The calibration shows that the system is not fully linear. The repeatability of poses in the same sense of rotation has an RMSE of 0.27° and a standard deviation below 0.3° for pitching and 4.7° for yawing averaged over three measurements. When comparing the same poses in clockwise and counter-clockwise direction the RMSE is 12.8° and 5.7° for pitching and yawing, respectively. This is likely due to motor hysteresis.

## 1 Problem

While robotic surgery is widely used in clinical practice [1], their lack of haptic feedback, soft tissue interaction or task automation remains challenging and represents active research topics [2]. Using a commercial device like the daVinci© System (Intuitive Surgical, Inc.) or the RAVEN II™ (Applied Dextricity) for such research is often not feasible, as the cost is substantial while flexibility and accuracy are too limited for the evaluation of new methods [3, 4]. Particularly, clinical master-slave systems typically rely on visual feedback and the complex and sometime wire driven kinematic design further complicates reproducible control of the instrument end-effector [3, 5].

In contrast, flexible and lightweight robot arms are widely available and simple to control, e.g., the Universal Robots UR3 and UR5 series or the KUKA LBR iiwa. Hence, we present a small, flexible wireless instrument adapter to mount EndoWrist© tools (Intuitive Surgical, Inc.) to robotic arms in order to facilitate and improve research on active topics. We present the design of the adapter, the control approach, and a calibration setup to study precise motion control despite the slack and coupling of the different degrees of freedom due to the wire driven mechanism inside the instruments.

## 2 Material und Methods

### 2.1 The EndoWrist© Instrument

All EndoWrist© Instruments are based upon the same mechanical principle: Four Bowden wires control a rotatable shaft and the specialized instrument placed at the tip of the same. The Bowden wires can be actuated via four discs with two cylindrical elevations at the bottom side of the Dock Assist Tool, representing one of the three rotatory degrees of freedom (Fig. 1). The discs can be actuated individually but there exist certain dependencies between three of the four discs. Disc 1 enables the rotation around the z-axis and is the only decoupled DOF. Disc 2 enables rotation around the x-axis, resulting in pitching of the tip. Disc 3 and 4 enable rotation around the y-axis, each controlling the yawing of one jaw side. The actuation of disc 2, 3 and 4 is coupled which results in a restriction of either range or requests simultaneous movement. In addition, tracking markers visible under ultraviolet light were added to the instruments tip as artificial landmarks. These landmarks facilitate the recognition of features, e.g., the pivot point, by a camera setup. Moreover, the minimal necessary force and holding torque for the disc actuation were measured with a force/torque sensor (SRI, M3703A) for motor dimensioning.

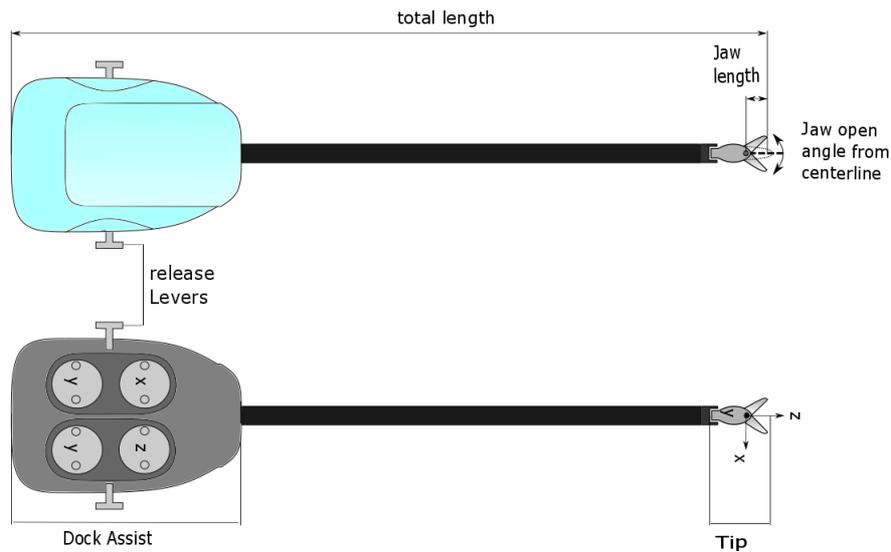

***Fig. 1:*** *The figure shows a schematic of the top (upper) and bottom (below) side of an EndoWrist© Instrument. All Bowden wires end in the Docking Assist Tool and are winded each onto a spool, which can be rotated by a disc around the respective axis x, y, or z (bottom view). The axes assign the rotational degrees of freedom of the instrument tip: Yawing around the y-axis, pitching around the x-axis and rolling around the z-axis. Furthermore, each instrument has its individual jaw length, jaw open angle and total length.*

## 2.2 Instrument Adapter

Our prototype of the instrument adapter consists of two modular parts. Part one is the motor box with detachable cover. The instrument can be inserted by a slide on/ slide off mechanism to the cover. Furthermore, the box contains four motors as well as a spring force operated lock home mechanism, which keeps the instrument firmly in place when inserted and which can be loosened by the release levers (Fig. 1). Simultaneously, this mechanism lifts the motors up, when the instrument slides in. Part two is the electronic box, which contains the motor controllers and a Bluetooth shield. In addition, the box leaves place for a standard size LiPo battery for autonomous power supply. The electronic box can be clipped to the motor box and can be further fixated with two screws. The box and the motor adapters placed on the motor shafts are 3D printed. The motor box measures 100 mm x100 mm x112 mm, the electronic box 150 mm x100 mm x135 mm. In total, the connected boxes measure 150 mm x100 mm x160 mm.

Four stepper motors (SOYO®, NEMA 11 SY28STH45-0674A) with a full step resolution of 1.8°/step were chosen for the discs' actuation. They permit to cover the full range of the instruments' motion without gearing and have sufficient torque. The motor control board is based on the Arduino platform using an ATmega2560 microprocessor (ATMEL), which was mounted to an additional shield (Tuandui™ Robotale, RAMPS 1.4 Arduino Mega Shield). The chip driver (Watterott electronic GmbH, Silent step Stick TCM2130 stepper motor driver) allows micro stepping up to1/256 step. For the measurements a micro step size of 1/16 micro step, which corresponds to 0.1125°/step, is set. The four end stops (Vishay® Semiconductors, TCST2103) are photo sensors, triggering a pull-up as soon as the contact between infrared emitter and phototransistor is interrupted. Beside the motor drivers and the end stops, a Bluetooth shield (Aukru®, HC-05 with Master RS232) is connected to the I/O-board shield, enabling wireless communication.

The library used for the motor control (Airspayce, AccelStepper 1.57) supports multiple simultaneous steppers and their respective speed and acceleration configuration. The main program contains an initialization sequence for the motors and for the inserted instrument in the setup loop and a sequence for the actual motor position input in the main loop. Commands can be passed wirelessly via the serial port, thus allowing other software, e.g., MATLAB to communicate with the microcontroller. When the controller gets the command to initialize, the motors begin to turn until the respective end stop is triggered, thus determining the initial position of the motor. Next, the program waits for the insertion of the instrument. When confirming the insertion, each motor turns along the full range of the respective disc in clockwise and counter-clockwise direction, so that the spring loaded motor-adapter-system snaps in. The motors return to the initial position, i.e. the center position (Fig. 2c)) and the controller waits for motor inputs.

## 2.3 Calibration Setup

### 2.3.1 Camera Calibration

A stereo camera system was set up to detect the displacement of the EndoWrist© Instrument (Fig. 2a)). This system consists of two cameras (Basler, acA1300 - 200uc) at a distance of 60mm. The working distance d is approximately 11mm. The cameras are arranged in an angle of approximately 30° to each other, i.e. 15° from center line, which provides two different perspectives of the recorded instrument tip and moves the instrument tip into the focus of the camera pair. To be able to recover depth from images, the camera parameters and orientation of the cameras coordinate systems to each other must be known. Therefore a stereo camera calibration using the camera calibration toolbox (Matlab) was performed. A calibration script was run, taking 50 pairs of images in order to improve the rotation and translation matrix. For error estimation, 100 image pairs were taken after the camera calibration and used to determine the camera RMSE.

### 2.3.2 Instrument calibration

In order to estimate starting values for the range of motion of the DOFs and to obtain transmission and coupling ratios an instrument calibration needs to be performed. Therefore, the instrument's dock Assist Tool is placed below the external sensor system showing the position of the discs at the bottom side. The discs are manually adjusted to every possible position maximum and their respective center position. The maxima were chosen as reference because they allowed a high reproducibility even by manual adjustment, as the discs have a clearly perceivable end stop they cannot pass. Moreover, it is possible to interpolate every other position from these twelve maxima. For each manually adjusted position, twenty pictures of the bottom side of the instrument are captured, showing the Dock Assist Tool disc arrangement. The images are then analyzed, calculating the averaged angle by which the disc has been rotated in reference to the horizontal centerline of the disc. The obtained values are equated to the motor ranges for the x, y1/y2 and z rotation. This analysis was performed by an automated MATLAB script. The same steps are followed in order to analyze the correspondent instrument tip position. Images are taken of the tip at the different maxima and then processed, profiting from the UV tracking markers, to assess the respective pitch, jaw and rotation range. These measured value pairs of motor input and corresponding instrument output were then linearly interpolated to estimate intermediate instrument positions.

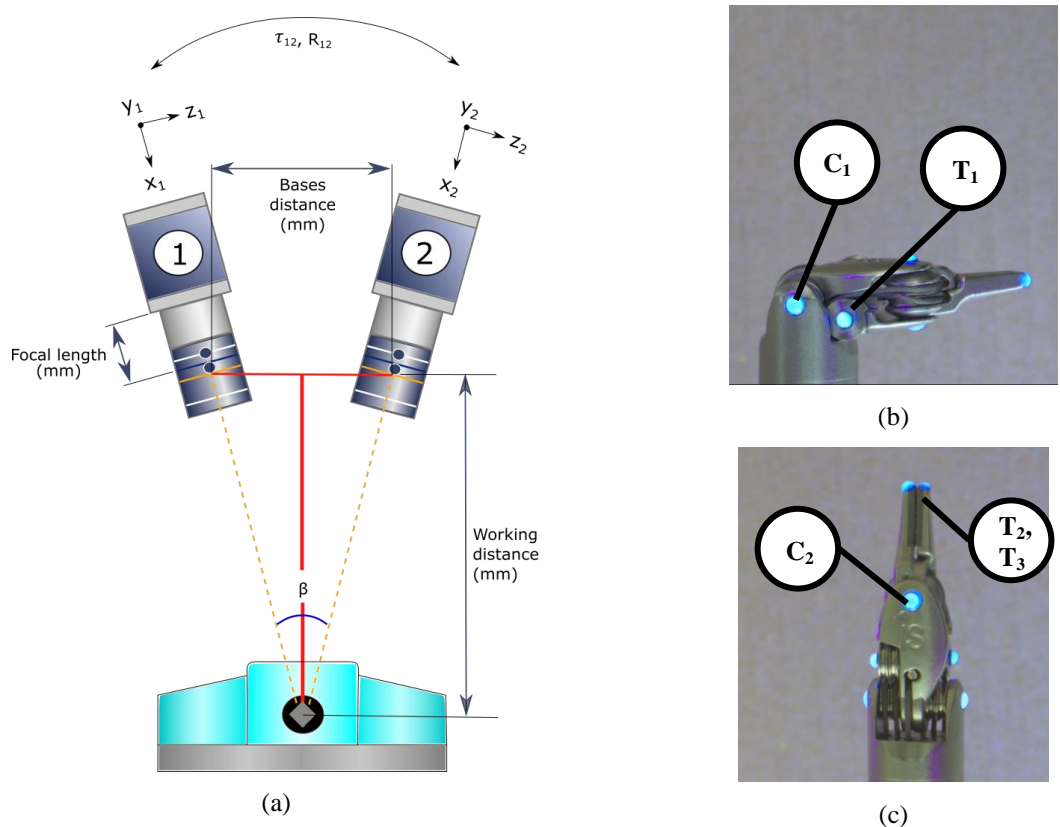

***Fig. 2:*** *a) Stereo camera setup. The base distance is 60mm, the focal length 51.5 mm, and the working distance 11mm. Rotation and translation are denoted by R12 and τ12, respectively. b) Side view of the instrument tip bent to the right, jaws closed middle with tracking marker C1 and T1 for pitch detection (blue). C) Top view of the instrument tip with tracking markers C2, T2 and T3 for Yaw detection (blue) in center position, jaws closed middle.*

### 2.3.3 System evaluation

For evaluation of the instrument adapter functionality, 20 different instrument poses are recorded and measured subsequently. For setup, the instrument is connected to the docking Station and its tip is placed in view of the sensor system, as described in section 2.3 and shown in Fig 2a). A custom-designed MATLAB routine connects to the cameras and the motor control and sends commands for image recording and motor movements. For pitching, five poses in clockwise and in counter-clockwise direction are recorded. For yawing, 25 poses in clockwise and counter-clockwise direction at each of the five poses for pitching were recorded. The tip is pitched from side to side in quarter steps. For yawing, one jaw side is opened and closed in quarter steps from fully open to fully closed, while the other jaw side remains fix at one of the five quarter positions. All position changes around the x-axis will be recorded from side view and all positon changes around the y-axis from top view, in order to have all tracking markers in sight (Fig. 2 b), c)). The measurement will be repeated three times, resulting in 3 x 500 images per camera. Measurement and influence of the rotational DOF are neglected in this experiment, as there exists no coupling to other DOFs. All captured images are analyzed in a sub routine, measuring the displacement while yawing and pitching respectively by means of the tracking marker. In order to estimate the reproducibility of each pose, the three different recordings for each pose in clockwise and counter-clockwise direction are averaged and the standard deviation is calculated. The mean values are then compared to the expectancy output value of the linear, theoretical system assessed in section 2.3.2 Additionally RMSE was calculated comparing clockwise and counter-clockwise direction to each other taking the clockwise poses as reference.

## 3     Results

### 3.1  Motor Test

The motor test resulted in a RMSE of 2.45% of the step size (N=3, n=482). The measured $torque_{motor,hold}$ is 0.035 Nm which is higher than the measured $torque_{hold,min}$, which is equal to 0.025 Nm.

### 3.2  Calibration Setup

The test of the stereo camera calibration resulted in an accuracy of 0.026 mm with a standard deviation of 0.0016 mm.

The recorded values, which were used for an initial calibration of the instrument, are presented in Tables 1 and 2. Table 1 shows the values measured while recording the Dock Assist Tool, Table 2 shows the values measured while recording the tip of *the Large Needle Driver*. By means of these measurements a set of possible motor position inputs and corresponding instrument position outputs were linearly interpolated, thus assessing an initial theoretical linear system with the transmission ratios and coupling ratios presented in Table 3.

|       | **Maxima CCW (°)** |      |      |        | **Center (°)** |      |      |        | **Maxima CW (°)** |      |      |        |
|-------|------|------|------|--------|------|------|------|--------|------|------|------|--------|
| **Z** | -167 |      |      |        | 0    |      |      |        | 167  |      |      |        |
| **X** | -85  |      |      |        | 0    |      |      |        | 85   |      |      |        |
|       | open | closed |    |        | open | closed |    |        | open | closed |    |        |
|       |      | left | mid  | right  |      | left | mid  | right  |      | left | mid  | right  |
| **Y₁** | -143 | 43  | -53  | -143   | -86  | 100  | 5    | -86    | -30  | 151  | 60   | -30    |
| **Y₂** | 21   | 21   | -69  | -156   | 76   | 76   | -15  | -102   | 128  | 128  | 39   | -48    |

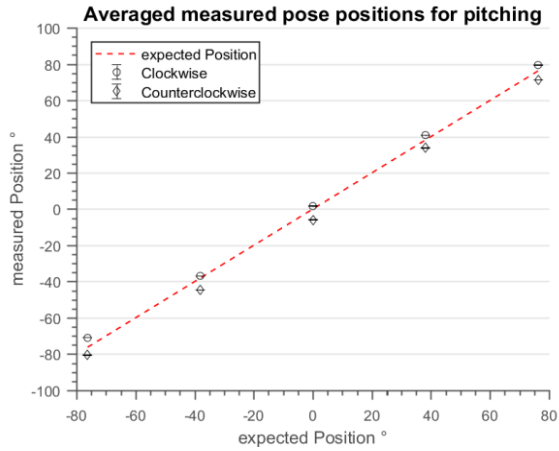

a)

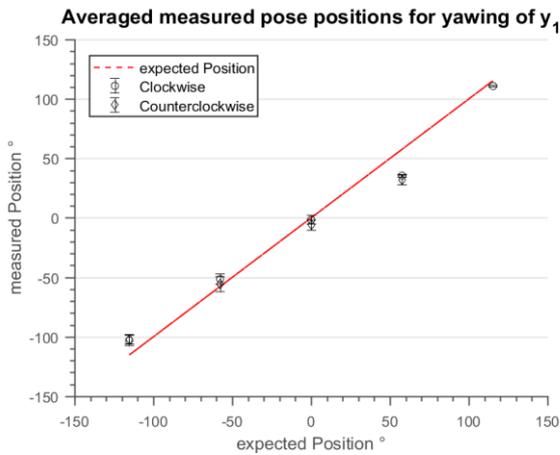

b)

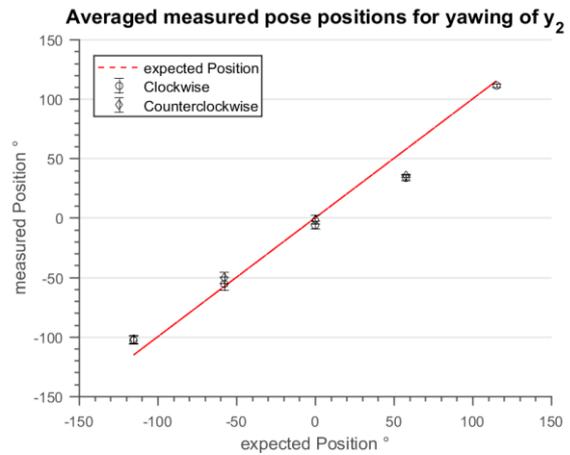

c)

***Fig. 3:*** *a, b and c show the results and their standard deviation of the system evaluation for pitching and yawing of $y_1$ and $y_1$ compared to the expected values.*

***Table 2:*** *Results for the maximal angles of the Large Needle Driver*

***Table 3:*** *Calculated transmission ratio from disc to tip and averaged coupling ratio for the Large Needle Driver*

|  | Maxima CCW ° | Center ° | Maxima CW ° |
|---|---|---|---|
| **Z** | -180 | 0 | 180 |
| **X** | -80 | 0 | 80 |
| **$Y_1$** | -115 | 0 | 115 |
| **$Y_2$** | -115 | 0 | 115 |

|  | Transmission ratio | Coupling Ratios | | | |
|---|---|---|---|---|---|
| **Z** | 1.1 | - | | | |
| **X** | 0.94 | $Y_1$ | 0.66 | $Y_2$ | 0.64 |
| **$Y_1$** | 1.3 | 1.52 | | | |
| **$Y_2$** | 1.3 | 1.57 | | | |

## 3.3  System Evaluation

Figures 3a, 3b and 3c show the mean position values averaged for three measurements for the pitching and yawing poses and the expected values from linear position interpolation. The standard deviation for clockwise and counter-clockwise direction respectively is ≤ 0.3° for pitching and up to 4.7° for yawing. The mean RMSE for both, clockwise and counter-clockwise direction amounts 12.7° for pitching and 5.7° for yawing. All necessary tracking markers have to be visible in both cameras. This is not the case for 196 images of the front view for each measurement, which are therefore excluded from analysis.

# 4　　Discussion

During instrument calibration, it could be observed that tools with the same instrument, e.g., the Large Needle Driver, have different ranges of motions for their discs. This implies that instrument calibration has to be performed for every provided tool. A possible explanation for this observation is that all provided instruments were second-hand and therefore already prestressed.

The results presented in Section 3.3 show that the system is not fully linear, comparing the interpolated values of the theoretical linear calibration to the values measured during the experiment. Nevertheless, the calibration resulted in a very good approximation for the pitching. Reproducibility is high for same direction whereas reproducibility comparing both directions is very low. This deflection is primarily caused by the motor back lash. Moreover, rounding error due to the fixed resolution of the motor steps lowered the accuracy of position information given to the motor. A smaller micro step size might increase the motor precision and lower hysteresis. For the y rotation, accuracy of the calibration is much lower. This could be due to the higher complexity and interdependency between the two jaw sides. The detection of the same point of the jaw tip tracking marker is less precise than for the other marker, as they are not circular. For further measurements the tracking system should be revised. The application of tracking markers already improved the calibration and made it more robust but at the same time the geometry of the markers complicates the detection of the same point in the two different camera perspectives, particularly when detecting the jaw tips (Fig. 2b)). Nevertheless, they are helpful compared to a mere geometrical tracking method and cause no impairment to the instrument as they can be easily removed.

# 5　　Conclusion

We presented a small, flexible and affordable adapter to use EndoWrist© instruments with general purpose robot arms. Our calibration setup and the adapter facilitate the evaluation of methods that require precise instrument control independently of a complete surgical robotics system. We demonstrate that the calibration results in a good initial parametrization, which can be improved successively by further measurements in order to achieve precise control of the instrument motion, e.g., to mimic grasping or tissue dissection.